\documentclass[10pt,twocolumn,letterpaper]{article}

\usepackage{cvpr}              
\definecolor{cvprblue}{rgb}{0.21,0.49,0.74}
\usepackage[pagebackref,breaklinks,colorlinks,allcolors=cvprblue]{hyperref}  

\usepackage[accsupp]{axessibility}  

\def\confName{CVPR}
\def\confYear{2026}

\title{Hierarchical Long Video Understanding with Audiovisual Entity Cohesion and Agentic Search}

\author{
    Xinlei Yin$^{1}$\thanks{Work done at Microsoft Research Asia.} \quad
    Xiulian Peng$^{2}\thanks{Corresponding authors.}$ \quad
    Xiao Li$^{2}$ \quad
    Zhiwei Xiong$^{1}$ \quad
    Yan Lu$^{2}\footnotemark[2]$\\
    $^{1}$University of Science and Technology of China \quad
    $^{2}$Microsoft Research Asia \\
    {\tt\small xyxl\_231829@mail.ustc.edu.cn, \{xipe, xili11, yanlu\}@microsoft.com, zwxiong@ustc.edu.cn}
}

\begin{document}
\maketitle
\begin{abstract}
Long video understanding presents significant challenges for vision-language models due to extremely long context windows.
Existing solutions relying on naive chunking strategies with retrieval-augmented generation, typically suffer from information fragmentation and a loss of global coherence.
We present HAVEN, a unified framework for long‑video understanding that enables coherent and comprehensive reasoning by integrating audiovisual entity cohesion and hierarchical video indexing with agentic search. 
First, we preserve semantic consistency by integrating entity-level representations across visual and auditory streams, while organizing content into a structured hierarchy spanning global summary, scene, segment, and entity levels. Then we employ an agentic search mechanism to enable dynamic retrieval and reasoning across these layers, facilitating coherent narrative reconstruction and fine-grained entity tracking.
Extensive experiments demonstrate that our method achieves good temporal coherence, entity consistency, and retrieval efficiency, establishing a new state-of-the-art with an overall accuracy of 84.1\% on LVBench. Notably, it achieves outstanding performance in the challenging reasoning category, reaching 80.1\%. These results highlight the effectiveness of structured, multimodal reasoning for comprehensive and context-consistent understanding of long-form videos.
\end{abstract}    
\section{Introduction}
\label{sec:intro}
The rapid growth of long-form video content in domains such as entertainment, education, and surveillance poses significant challenges for automated understanding systems.
Unlike short clips, hour-long videos demand reasoning over tens of thousands of frames and audio streams, where events unfold across extended durations and depend on subtle cross-scene relationships.
The interleaving of multiple entities, dynamic interactions, and evolving multimodal signals makes coherent interpretation and entity tracking particularly difficult.

\begin{figure}[t]
  \centering
   \includegraphics[width=0.99\linewidth]{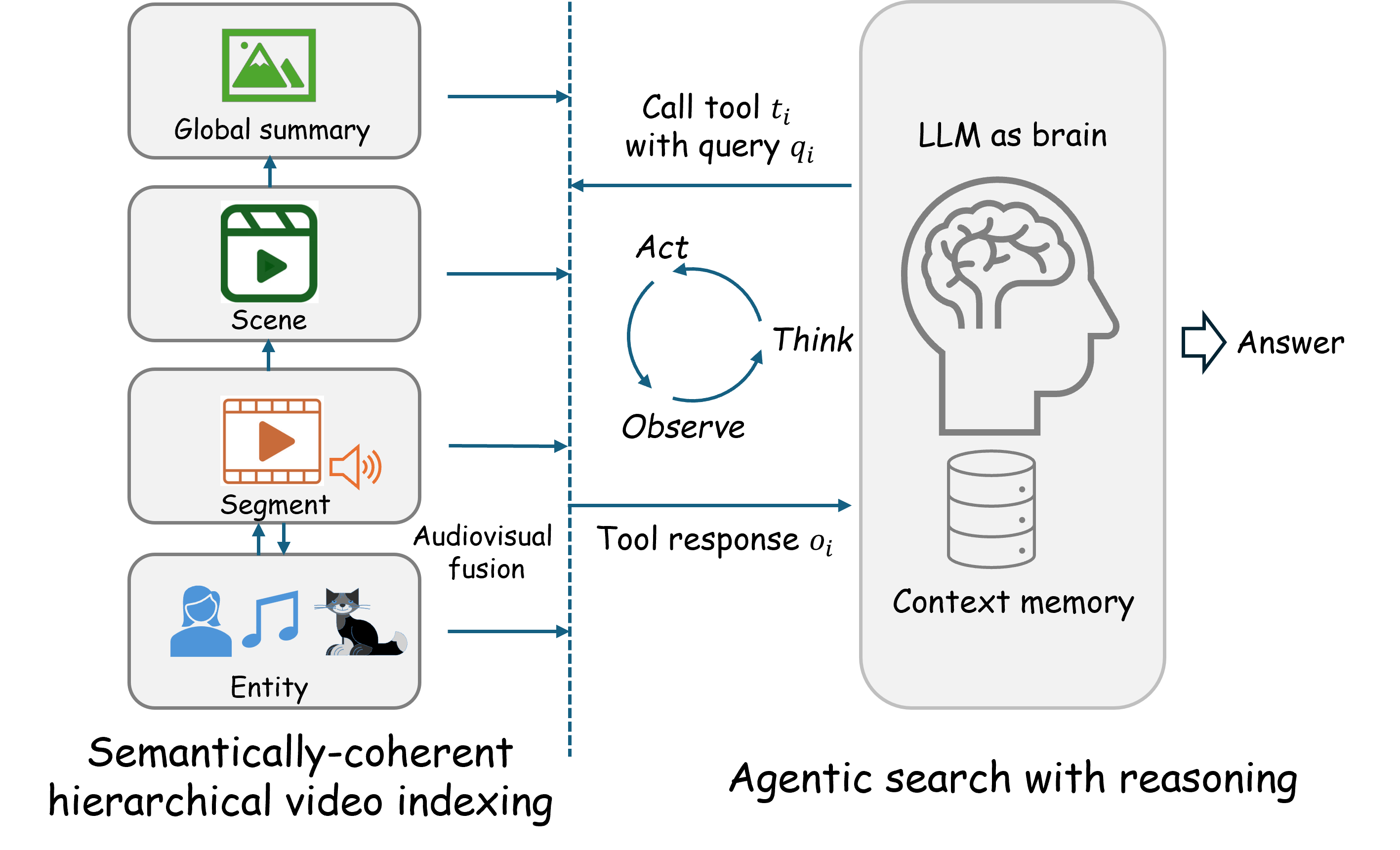}
   \caption{The proposed hierarchical video indexing with audiovisual entity cohesion and agentic search.}
   \label{fig:videographrag}
\end{figure}

Recent advances in large Vision-Language Models (VLMs) have led to substantial progress on short-video tasks such as captioning, question answering, and temporal grounding, yet their application to long videos remains constrained by limited context windows and computational bottlenecks.
To mitigate these issues, techniques such as adaptive sampling~\cite{Kai-Frame, QueryFS, BOLT}, and token compression~\cite{AdaReTake, MovieChat, Video-XL, AdaCM2} are proposed, enabling longer sequences under resource limits. Memory-based approaches~\cite{ReWind, MA-LMM} further extend temporal coverage by dynamically retaining and updating salient information. While these techniques improve scalability, they often sacrifice critical details or contexts, and struggle to preserve the semantic continuity across distant segments.

To address these issues, recent methods increasingly adopt retrieval-augmented generation (RAG) \cite{VideoRAG, luo2024videorag, videorag-ACL, AdaVideoRAG, shen2025vgent} to dynamically fetch relevant video segments, alongside agentic frameworks \cite{VideoAgent, LVAgent, zhang2025dvd} that autonomously plan and reason over the video content. However, both paradigms exhibit fundamental limitations in long-video scenarios.
First, retrieval is typically driven by isolated signals (e.g., clip-level captions), which yields fragmented or redundant evidence and severely weakens global narrative coherence.
Second, the absence of a hierarchical video representation deprives agents of the structural context needed for multi-level reasoning. Consequently, models can only resort to inefficient, multi-round retrievals to recover cross-segment continuity, introducing unnecessary complexity.

In this paper, we propose a unified agentic framework that shifts long-video understanding from fragmented retrieval to coherent, structured comprehension (see Fig.~\ref{fig:videographrag}).
Our approach is based on two core innovations: audiovisual entity cohesion and hierarchical indexing.
First, we introduce audiovisual entity cohesion, which leverages complementary audio and visual evidence to consolidate fragmented observations into consistent entities and enrich them with multimodal attributes. Particularly, we explicitly exploit speaker identity cues as a powerful yet often over-looked signal for entity consolidation, serving as an effective “glue” for maintaining long-range entity coherence. The resulting entity representations provide reliable building blocks for higher-level scene interpretation as well. 
Second, we construct a hierarchical database that organizes video content at multiple granularities, i.e. global summary, scene, segment, and entity, enabling flexible and multi-level retrieval. 
Built atop this hierarchy, our agentic search mechanism supports goal-driven reasoning across granularities. 
Ultimately, by integrating these components, our framework achieves scalable and holistic long-video understanding. Our contributions are summarized as follows:

\begin{itemize}
\item \textbf{Audiovisual entity cohesion}: we propose a cross-modal entity consolidation mechanism that maintains semantic consistency across time and modalities, effectively improving entity continuity and narrative coherence.
\item \textbf{Hierarchical indexing with agentic search}: we design a hierarchical indexing database with an agentic search strategy that dynamically navigates and reasons over this hierarchy, enabling efficient and structured information access.
\item \textbf{State-of-the-art performance}: we evaluate our framework on several long video understanding benchmarks, achieving superior performance over existing baselines, with an overall accuracy of 84.1\% on LVBench.
\end{itemize}

\section{Related Work}
\label{sec:related}

\paragraph{Long Video Understanding with large VLMs}
Large vision-language models have been extended to long-form video tasks using techniques such as adaptive sampling \cite{Kai-Frame, QueryFS, BOLT} and token compression \cite{AdaReTake, MovieChat, Video-XL, AdaCM2}, which reduce token overhead by selecting salient frames or merging redundant information across time and modality. These strategies enable longer sequence processing under memory and compute constraints but often sacrifice critical details or incur additional online computation. Memory-based approaches \cite{ReWind, MA-LMM} dynamically retain and update salient content, extending the temporal receptive field and supporting reasoning over longer durations. In addition, other approaches such as LongVLM \cite{LongVLM} and VideoStreaming \cite{VideoStreaming} leverage hierarchical token aggregation and memory propagation with global-local semantics to support scalable long video understanding. Despite these advances, existing solutions struggle to balance global coherence with local detail, making it difficult to reason across distant scenes or maintain entity continuity over time. 

\paragraph{RAG-based Long Video Understanding}
RAG has emerged as a promising paradigm for scaling long video understanding by segmenting videos into retrievable units and dynamically fetching relevant context during inference.  Recent works have advanced this paradigm in several directions, including aligned video-text chunking \cite{videorag-ACL}, graph-based entity semantics \cite{VideoRAG}, omni-contextual adaptive selection \cite{AdaVideoRAG}, and temporal dependency modeling \cite{shen2025vgent}. However, their reliance on fragmented segments and lack of global context limit temporal coherence, thus making complex reasoning tasks challenging.

\paragraph{Long Video Agents}
Unlike static RAG-based retrieval, recent work leverages LLMs as autonomous agents for iterative planning, retrieval, and reasoning over long videos, enabling dynamic interaction through tool use and structured search. Representative works include VideoAgent \cite{VideoAgent}, which treats LLMs as agents that iteratively retrieve and interpret video segments; VideoTree \cite{wang2024videotree}, which introduces a tree-based representation for adaptive exploration; LVAgent \cite{LVAgent}, which coordinates multiple agents in multi-round collaboration; and Deep Video Discovery (DVD) \cite{zhang2025dvd}, which supports tool-augmented search over global and local video content. Additionally, DrVideo \cite{DrVideo} reformulates long videos into document-like structures for iterative retrieval-augmented inference. While these frameworks improve interactive reasoning and scalability, their underlying databases remain simple (e.g., frames, clip captions, visual entities) and often require heavily iterative processes to find the answer, leading to high computational costs.

\paragraph{Hierarchical Video Representation}

Long video understanding often requires modeling local-global context and hierarchical relationships across scenes, segments, and entities. Prior works address this via local-global aggregation~\cite{LongVLM, VideoStreaming}, tree-based structure~\cite{wang2024videotree}, and graph-based retrieval~\cite{AdaVideoRAG, VideoRAG}, while \cite{zhang2025dvd} tracks subjects in a global registry. Although these methods improve semantic coherence, some models incur high online computation and lack a unified offline hierarchical index spanning video, scene, segment, and entity levels, which is a core feature of our framework for efficient and coherent long video reasoning.

\begin{figure*}[t] 
  \centering
  \includegraphics[width=\textwidth]{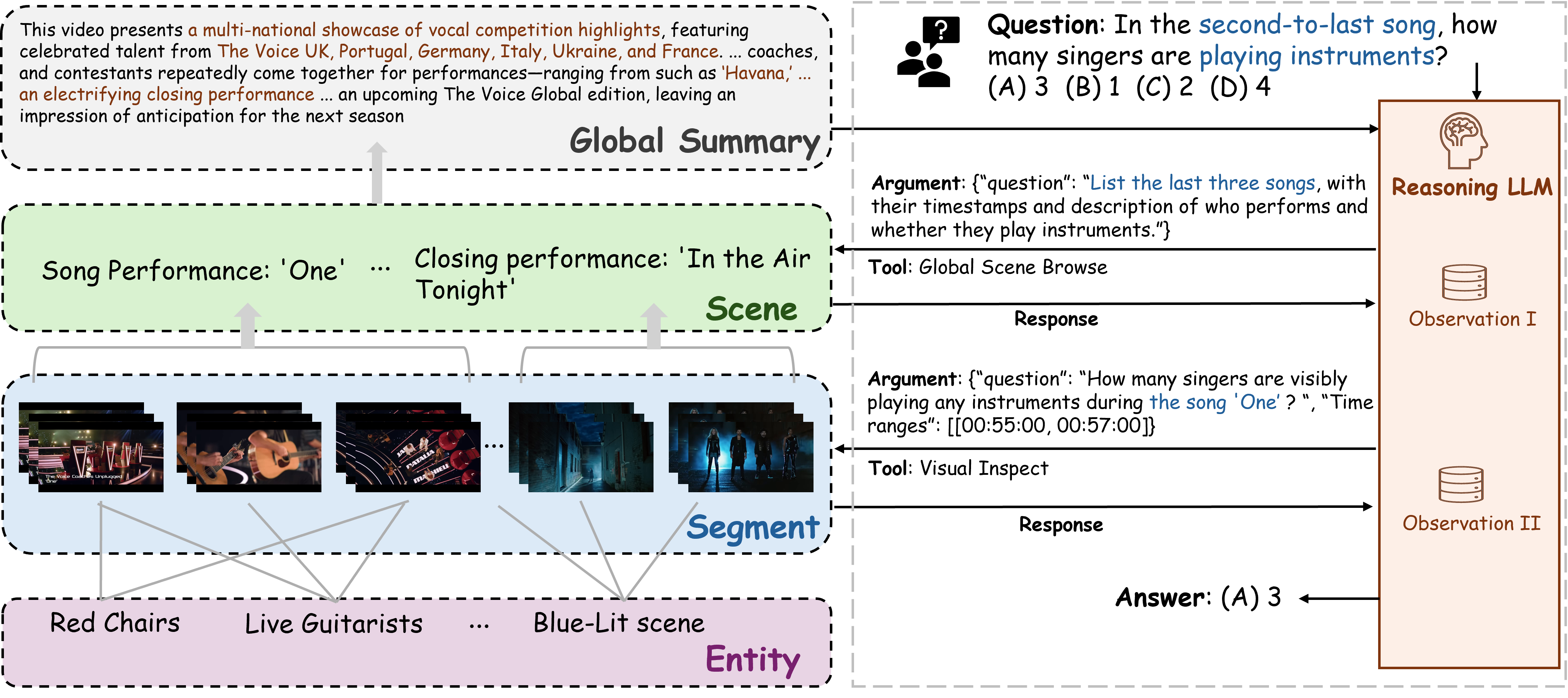}
  \caption{Overview of our framework. Left: hierarchical database. Right: agentic reasoning. The reasoning LLM calls tools iteratively to collect information and answer the question.}
  \label{fig:case_pipeline}
\end{figure*}

\section{Method}
\label{sec:proposed}
\subsection{Overview}
Recent RAG and agentic frameworks typically construct databases from clip-level captions~\cite{zhang2025dvd} and global entity sets that link disjoint clips \cite{VideoRAG, AdaVideoRAG,shen2025vgent}. However, this combination of local-clip and global-entity does not capture the inherent hierarchical nature of video content, where entities, events, and scenes evolve at different temporal scales and interact over long horizons. Consequently, both global/scene-level queries (e.g. \textit{``what is the video about?''} or \textit{``What is the name of the song sung by the third contestant?''}) and locally ambiguous queries (e.g., \textit{``what does the protagonist do at 12{:}00-12{:}30?''}) require long-range contextual reasoning. Relying solely on online retrieval of local fragments can yield incomplete evidence or overwhelm the model with redundant and incoherent information.

To address these issues, we propose a framework for \textit{semantic-coherent hierarchical video indexing and agentic retrieval} that performs offline parsing to construct a structured database and enables query-dependent navigation over multiple granularities. A key ingredient is \textit{audiovisual entity cohesion}: rather than treating audio as subtitle text only, we leverage speaker identity as a complementary and often more stable cue for entity continuity. Speaker identity can remain informative even when visual evidence becomes unreliable due to occlusion, changes in viewpoint or lighting, motion blur, crowded scenes, shot transitions, or off-screen speakers. This cross-modal cue helps consolidate fragmented entity observations and provides more reliable building blocks for higher-level scene interpretation. 

Concretely, we construct a four-level hierarchical database
\begin{equation}
\mathcal{D} = \{ \tilde{\mathcal{C}}, \tilde{\mathcal{E}}, \tilde{\mathcal{S}}, \tilde{\mathcal{G}} \},
\end{equation}
where $\tilde{\mathcal{C}}$, $\tilde{\mathcal{E}}$, $\tilde{\mathcal{S}}$, and $\tilde{\mathcal{G}}$ denote segment-level audiovisual information, canonical audiovisual entities, scene summaries, and a global summary, respectively (Fig.~\ref{fig:videographrag}). We leverage speaker diarization to maintain consistent speaker identities across the entire video, which are used together with visual information to support robust entity consolidation. Meanwhile, a LLM-based temporal abstraction pipeline merges semantically related segments into scene-level and global summaries, providing higher-level anchors for long-range reasoning.

During inference, our agent performs \textit{query-dependent adaptive search} over $\mathcal{D}$ via a $\textit{think-act-observe}$ loop (Fig.~\ref{fig:videographrag}), following~\cite{zhang2025dvd}. A suite of multi-granularity retrieval tools allows the agent to dynamically navigate at multiple levels, enabling efficient evidence collection and accurate answer generation for diverse query types.

\subsection{Database Construction}
\subsubsection{Audio Information Extraction}
Audio provides complementary cues for long-video understanding, including time-aligned transcripts and speaker identity. Beyond using transcripts/subtitles as auxiliary text, we incorporate speaker identity as a long-range consistency signal that remains informative even when visual cues degrade (e.g., occlusion, shot transitions, view changes, or off-screen speakers). This is particularly valuable for dialog-driven content such as documentaries, TV shows, and vlogs. We employ automatic speech recognition (ASR) and speaker diarization using WhisperX~\cite{bain2023whisperx}, which jointly produce accurate transcripts with timestamps and consistent speaker labels. These annotations enable us to understand not only \textit{``what was said and when''} but also \textit{``who said it,''} a key factor for maintaining entity-level coherence across time. 

\subsubsection{Segment Information Extraction}
We uniformly divide the video into fixed-length temporal segments as the basic unit of local evidence. For each segment $i$, we extract the audio annotations, including speaker labels $P_i$ and timestamped transcripts $T_i$. These audio annotations, together with sampled video frames, are provided to a VLM to produce a segment-level caption $V_i$ and a speaker-aware description $P_i'$, which associates each speaker label with salient visual cues in the segment (e.g., appearance, actions, role cues when available). We then construct the segment textual representation $C_i^t$ by
\begin{equation}
C_i^t = [P_i' \, ; \, T_i \, ; \, V_i].
\end{equation}
This representation grounds \textit{who said what and when} in a localized visual context and serves as the basis for subsequent entity consolidation and cross-temporal reasoning.

Since captioning may miss fine-grained visual details, we further augment each segment with a visual embedding $C_i^v$ using the multimodal retrieval model UNITE~\cite{unite}. The final segment representation is 
\begin{equation}
C_i=[C_i^t \, ; \, C_i^v],
\end{equation}
which composes the segment database $\tilde{\mathcal{C}}$. In our implementation, each segment spans 30 seconds and we sample 20 frames per segment for caption generation.

\begin{figure*}[ht] 
  \centering
  \includegraphics[width=0.9\textwidth]{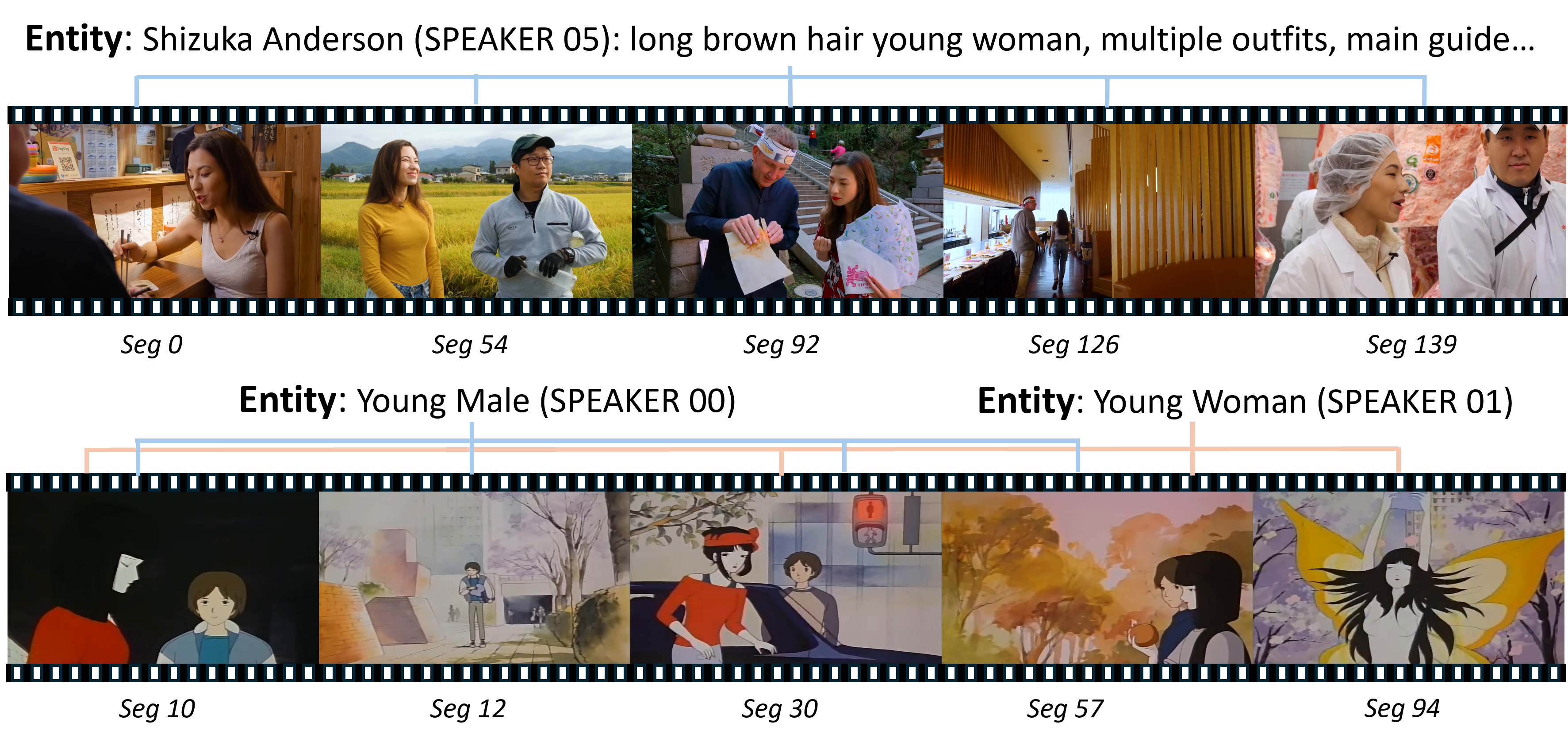}
  \caption{Speaker identity for entity consolidation. Segments with long temporal gaps can still be linked to the same speaker entity through a shared speaker identity.}
  \label{fig:spk_cases}
\end{figure*}

\subsubsection{Audiovisual Entity Extraction}
For each segment $i$, we employ an LLM to extract salient entities from the textual representation $C_i^t$, covering characters, locations, and events. This yields
\begin{equation}
E_i = \{ e_1^i, e_2^i, \ldots, e_{N_i}^i \},
\end{equation}
where each $e_k^i$ includes an entity name and a concise description derived from the segment’s audiovisual context.

Consolidating these entities across long videos is non-trivial. Simple heuristics such as name matching or embedding-similarity thresholding~\cite{VideoRAG, shen2025vgent} can be brittle: the same character may be split across segments due to appearance or viewpoint changes, whereas distinct but visually similar characters may be erroneously merged. To address this, we perform entity consolidation in two stages, \textit{embedding-based clustering} followed by \textit{LLM-based canonicalization}. First, we embed each entity description using a text encoder $f_{\text{text}}(\cdot)$, $z_k^i = f_{\text{text}}(e_k^i)$, and cluster the embeddings to form candidate groups of cross-segment correspondences. Second, an LLM revisits each cluster and either produces a canonical entity summary or splits the cluster into subgroups when semantic conflicts are detected. This verification step mitigates both over-merging and over-splitting, yielding a refined set of canonical entities
\begin{equation}
\tilde{\mathcal{E}}_g = \{ \tilde{e}_1, \tilde{e}_2, \ldots, \tilde{e}_J \}.
\end{equation}

To further strengthen long-range coherence, we incorporate speaker identity cues into consolidation. When multiple segments share the same diarized speaker label, we treat this as a strong consistency signal and prioritize merging the corresponding character-related entity mentions, even when visual or textual descriptions vary due to shot changes, occlusion, or other degradations. This yields robust audiovisual entity cohesion and improves identity continuity across the video. Figure~\ref{fig:spk_cases} illustrates two examples in which characters undergoing dramatic appearance changes are correctly consolidated using speaker identity cues. Such consolidation is crucial for questions like \textit{``How does the emotion change on Sarah's face when interviewed?''}, where speaker identity links disjoint fragments to the identity \textit{``Sarah''}, capturing a long-range continuity that isolated clip captions fail to provide.

After consolidation, each canonical entity $\tilde{e}_j$ is associated with a global description and a set of linked segments $Q_j$. Directly incorporating all segments linked to the top-$K_1$ entities can be expensive and may introduce query-irrelevant noise. We therefore perform \textit{entity-centric re-captioning} during offline construction, producing a focused description \(\tilde{C}^t_{i,j}\) centering on the entity $\tilde{e}_j$ for each linked segment \(i \in Q_j\). The final entity database contains both canonical entities and fine-grained entity-segment descriptions, $\tilde{\mathcal{E}}=\{\tilde{\mathcal{E}}_g \, ; \, \tilde{\mathcal{E}}_e\}$, where $\tilde{\mathcal{E}}_e=\{\tilde{C}^t_{i,j}|j=1,2,\ldots,J, i \in Q_j\}$. More details on this process are provided in Sec.~\ref{supp:entity-centric} of the supplementary material.

\subsubsection{Scene Segmentation and Global Summary}
Long videos often consist of multiple temporally extended scenes with coherent narratives, recurring entities, and consistent environments. Existing approaches typically segment videos into short fixed-length clips (e.g., 5 seconds in~\cite{zhang2025dvd}) or aggregate fixed-size chunks~\cite{VideoRAG}, which can capture local moments but may miss long-range temporal structure. We therefore perform adaptive \textit{scene-level aggregation} based on segment descriptions $\{C_i^t\}_{i=1}^N$, where $N$ is the total number of segments. We split the sequence into overlapping chunks and use an LLM to group consecutive and semantically related segments into scenes:
\begin{equation}
    \mathcal{S} = \{ s_j \}_{j=1}^M, \quad s_j = \{ c_{a_j}, c_{a_j+1}, \ldots, c_{b_j} \},
\end{equation}
where each $s_j$ represents a temporally contiguous scene with a consistent narrative focus, and $c_i$ denotes the $i$-th segment. The boundaries $(a_j, b_j)$ are adaptively determined by the LLM based on the semantic continuity.

Each scene $s_j$ is then summarized by an LLM to produce a concise scene-level description $\tilde{s}_j$, capturing its key characters, events, and transitions. The collection of these summaries forms the high-level scene set $\tilde{\mathcal{S}} = \{ \tilde{s}_1, \tilde{s}_2, \ldots, \tilde{s}_M \}$. Finally, a global summary $\tilde{\mathcal{G}}$ is generated from $\tilde{\mathcal{S}}$, describing the main storyline, recurring entities, and overall context (e.g., background and video type).

\subsection{Agentic Search with Reasoning}
Similar to~\cite{zhang2025dvd}, our agentic retrieval adopts an iterative \textit{think-act-observe} loop for adaptive information retrieval. At the core, we use a reasoning LLM as the planner, which formulates intermediate queries (\textit{think}), invokes specialized tools with appropriate parameters (\textit{act}), and incorporates tool outputs into subsequent reasoning (\textit{observe}). This synergy between LLM-driven reasoning and tool interaction enables progressive query refinement, evidence-guided navigation, and accurate answer generation. We initialize the agent with the global summary $\tilde{\mathcal{G}}$, and enrich its context with tool outputs across multiple iterations. 

\subsubsection{Multi-Granularity Tools}
To support adaptive reasoning over long videos, we equip the agent with a suite of multi-granularity retrieval and inspection tools operating on different levels of the hierarchical database $D$. We denote the toolset as
\begin{equation}
    \mathcal{T} = \{ T_{\text{scene}}, T_{\text{caption}}, T_{\text{visual}}, T_{\text{entity}}, T_{\text{inspect}} \}.
\end{equation}
Each tool $T_i \in \mathcal{T}$ is a callable function that takes a textual query $q$ with optional time ranges $\tau = [t_s, t_e]$ and returns a textual response $r$ with associated timestamps $\tau'$, i.e., $T_i(q,\tau;D) \rightarrow (r, \tau')$.

The \textbf{Global Scene Browse} tool $T_{\text{scene}}$ performs coarse navigation over scene summaries $\tilde{\mathcal{S}}$ to localize relevant scenes and time ranges. The \textbf{Segment Caption Search} $T_{\text{caption}}$ retrieves fine-grained evidence from segment descriptions in $\tilde{\mathcal{C}}$ via text embedding matching. The \textbf{Segment Visual Search} $T_{\text{visual}}$ complements caption retrieval with visual-semantic search using cross-modal embeddings by UNITE~\cite{unite} to capture cues missing from text. The \textbf{Entity Search} $T_{\text{entity}}$ conducts entity-centric retrieval over canonical entities $\tilde{\mathcal{E}}_g$ and their linked evidence $\tilde{\mathcal{E}}_e$ to gather long-range information about specific entities; and the \textbf{Inspection} tool $T_{\text{inspect}}$ provides localized inspection within specified time spans and consists of two complementary modules: \textit{Clip Caption Inspect} ($T_{\text{inspect}}^{\text{tex}}$), which examines textual descriptions to identify what occurred during the interval, and \textit{Visual Inspect} ($T_{\text{inspect}}^{\text{vis}}$), which performs VLM-based visual verification for fine-grained confirmation.

During inference, the agent dynamically selects and composes these tools based on the query intent and current context, prioritizing low-cost text retrieval before higher-cost visual inspection. Detailed tool definitions and implementation are provided in Sec.~\ref{supp:tools} of the supplementary material.

\subsubsection{Multi-Step Reasoning}
In addition to $(\mathcal{T}, D)$, the agent maintains a context memory $\mathcal{M}$, initialized with the global summary $\tilde{\mathcal{G}}$, i.e. $\mathcal{M}_0=\{\tilde{\mathcal{G}}\}$. Let $q_{org}$ denote the original question. At reasoning step $t$, the planner dynamically selects to call a tool $T_{a_t}$ and forms an intermediate query $q_t$ with an optional time window based on the current memory: $a_t = \pi_{\text{LLM}}(q_{org}, \mathcal{T}, \mathcal{M}_t)$, where $\pi_{\text{LLM}}$ denotes the LLM-driven policy that determines which tool to invoke. The tool results $(r_t, \tau'_t) = T_{a_t}(q_t;D)$, and the tool call context are integrated into the context memory to guide subsequent reasoning, i.e. $\mathcal{M}_t=\{\mathcal{M}_{t-1};(T_{a_t}, q_t);(r_t, \tau'_t)\}$. The process repeats until it finds the answer or a maximum number of steps is reached. Figure~\ref{fig:case_pipeline} shows an example for a counting-like query that needs long contexts. The agent first calls the \textit{Global Scene Browse} tool to locate the \textit{``second-to-last song''} and obtain its content and time span. It then calls \textit{Visual Inspect} tool to analyze the segment of the \textit{song ``One'' (55:00-57:00)}, which returns \textit{``two singers playing guitar and one playing percussion''}, enabling the agent to derive the final answer.

\begin{table*}[ht]
\small
  \caption{Comparison on LVBench. Accuracy (\%) is reported. Bold indicates the best performance, and underscore denotes the second-best.}
  \vspace{-3pt}
  \label{tab:lvbench}
  \centering
  \begin{tabular}{@{}lccccccc@{}}
    \toprule
    Methods & ER & EU & KIR & TG & Rea & Sum & Overall \\
    \midrule
    \textit{Proprietary VLMs} \\
    GPT-4o \cite{GPT-4} & 48.9 & 49.5 & 48.1 & 40.9 & 50.3 & 50.0 & 48.9 \\
    OpenAI o3 \cite{O3} & 57.6 & 56.4 & 62.9 & 46.8 & 50.8 & 67.2 & 57.1 \\
    \midrule
    \textit{Open-source VLMs} \\
    Qwen2.5-VL-72B \cite{Qwen2.5-VL} &- &- &- &- &- &- & 47.7 \\
    AdaReTake \cite{AdaReTake} & 53.0 & 50.7 & 62.2 & 45.5 & 54.7 & 37.9 & 53.3 \\
    Seed1.5-VL-Thinking-200B \cite{Seed1.5-VL} & 65.4 & 63.4 & 68.0 & 53.6 & 63.7 & 46.6 & 64.6 \\
    \midrule 
    \textit{RAG and agents} \\
    VideoRAG \cite{VideoRAG} & 47.4 & 49.3 & 57.1 & 36.5 & 43.9  & 39.7 & 49.2 \\
    VideoTree \cite{wang2024videotree} & 30.3 & 25.1 & 26.5 & 27.7 & 31.9 & 25.5 &28.8 \\
    VideoAgent \cite{VideoAgent} &28.0 &30.3 &28.0 &29.3 &28.0 &36.4 &29.3 \\ 
    VideoLucy \cite{zuo2025videolucy} & 54.3 & 59.8 & 75.6 & 51.7 & 55.9 & 49.1 & 58.8 \\
    DVD \cite{zhang2025dvd} &73.4 &73.3 &80.4 &72.3 &70.7 &74.1 &74.2 \\
    DVD w. subtitle \cite{zhang2025dvd} &75.5 &77.1 &79.0 &72.7 &68.7 &\textbf{84.5} &76.0 \\
    \midrule
    Ours & \underline{79.9}&\underline{82.6} &\underline{84.0} &\textbf{86.7} &\underline{79.6} &\underline{79.3} & \underline{81.0} \\
    Ours (2 fps) & \textbf{83.2}& \textbf{84.8} &\textbf{88.2} &\underline{82.0} &\textbf{80.1} &\textbf{84.5} &\textbf{84.1} \\
    \bottomrule
  \end{tabular}
\end{table*}

\section{Experimental Results}
\label{sec:results}

\begin{table}[ht]
\small
  \caption{Comparison on other long video benchmarks. We use 0.67 fps for captioning in our approach. “Video-MME (L)” denotes the long split of Video-MME. “LVB” denotes LongVideoBench.}
  \label{tab:benchmarkmore}
  \centering
  \begin{tabular}{@{}lcccc@{}}
    \toprule
    Methods  & \multicolumn{2}{c}{Video-MME (L)} & LVB & EgoSchema  \\
             & w/o sub & w sub & Long & Val  \\
    \midrule
    GPT-4o & 65.3 &72.1 &60.9 &70.4 \\
    OpenAI o3 &64.7 &- &60.6 &63.2  \\
    \midrule
    Qwen2.5-VL-72B & 63.9 &- &- &- \\
    AdaReTake &65.0 &76.4 &-  &- \\
    \midrule
    VideoTree  &- &- &- &67.0 \\
    VideoAgent &- &- &- &63.2 \\    
    VideoLucy  & 66.8 & - & -&- \\
    DVD & 67.3 &- & 68.6 &76.6 \\
    \midrule
    Ours &- &\textbf{82.8} &\textbf{78.2} &\textbf{81.6} \\
    \bottomrule
  \end{tabular}
\end{table}

\subsection{Benchmark and Implementation Details}

\noindent\textbf{Benchmark}
We evaluate our approach on four widely-used long video understanding benchmarks, including LVBench~\cite{LVBench}, Video-MME \cite{VideoMME}, LongVideoBench \cite{LongVideoBench} and EgoSchema \cite{EgoSchema}. The LVBench contains 1,549 questions across 103 videos, with an average duration of 4,101 seconds. The dataset covers six diverse categories: temporal grounding (TG), summarization (Sum), reasoning (Rea), entity recognition (ER), event understanding (EU), and key information retrieval (KIR). With high-quality ground-truth annotations, LVBench provides a reliable and challenging benchmark for evaluating comprehensive long video understanding capabilities. For Video-MME, we evaluate it on its \textit{long} split, which includes 300 videos and 900 questions, with durations ranging from 30 to 60 minutes. For LongVideoBench, we focus on the \textit{long} subset in the \textit{val} split, containing 188 videos and 564 questions, with durations between 900 and 3600 seconds. Most videos in this subset lack audio tracks and we utilize the official subtitles without speaker identity. EgoSchema serves as a diagnostic benchmark for long-form understanding and reasoning. We evaluate on its \textit{val} split, which contains 500 videos and 500 questions, with each video lasting three minutes and no audio provided. For simplicity, we use audio streams only when the language is English across all datasets.

\noindent\textbf{Implementation Details}
For database construction, we employ GPT-4.1~\cite{GPT-4} to generate segment-level captions and summarize scenes and entities. Each 30-second segment is captioned using 20 sampled frames (0.67 fps) to ensure efficiency while maintaining coverage. During agentic search, we use OpenAI o3~\cite{O3} as the reasoning planner, which aggregates information from tool outputs and produces answers, with a maximum reasoning depth of 10 steps. For query-aware re-captioning during tool calls, we sample 30 frames per segment (1 fps) to provide sufficient visual context. Additionally, we leverage OpenAI o3 as the VLM used in \textit{Visual Inspect} tool and set its maximum number of input frames to 50. Unless otherwise specified, the same settings are applied across all datasets.

\subsection{Comparison with State-of-the-Art}
We compare our framework against several leading approaches for long video understanding, including proprietary large VLMs~\cite{GPT-4,O3}, open-source VLMs~\cite{Qwen2.5-VL, AdaReTake, Seed1.5-VL}, RAG-based systems~\cite{VideoRAG}, and video agents~\cite{wang2024videotree,VideoAgent,zhang2025dvd,zuo2025videolucy}. We reproduce the results of VideoRAG~\cite{VideoRAG} on LVBench using the official implementation, while other baseline results are taken directly from published reports.

Table \ref{tab:lvbench} summarizes the results on LVBench. Our approach consistently outperforms all baselines in overall accuracy, with especially strong gains in reasoning (Rea), a challenging category for existing methods, and temporal grounding (TG). These improvements highlight the effectiveness of our semantically consistent hierarchy coupled with multi-level agentic search. In entity recognition (ER), our method also surpasses DVD, which relies on global subject registries, demonstrating the robustness of our audiovisual entity cohesion strategy. Notably, our system achieves these results using much coarser temporal segmentation (30-second segments with 20 frames for captioning) and fewer reasoning iterations (up to 10) compared to DVD’s finer-grained setup (5-second segments at 2 fps) and deeper reasoning depth (up to 15 steps), underscoring both the efficiency and scalability of our approach. Further increasing the sampling rate to 2 fps for each 30-second segment (denoted as \textit{Ours (2 fps)} in Table \ref{tab:lvbench}) yields an additional 3.1\% accuracy improvement, attributed to the richer textual descriptions enabled by denser captioning.

Table \ref{tab:benchmarkmore} further presents the comparison results on other benchmarks. Our method achieves the highest scores on all benchmarks, reaching 82.8\% on Video-MME (long), 78.2\% on \textit{long} \textit{val} split of LongVideoBench, and 81.6\% on EgoSchema, even though most videos in LongVideoBench and EgoSchema lack audio streams. These results underscore the effectiveness of reasoning across the proposed hierarchical structure with semantic coherence. 

\subsection{Ablation Study}
\subsubsection{Ablation on Various Modules}
We conduct an ablation study on LVBench to evaluate the contributions of several key components: the \emph{Speaker} clues and \emph{Transcript} in audio stream, the \emph{Hierarchical} organization of video semantics, and the \emph{Visual embed} for visual retrieval. We compare four variants in Table~\ref{tab:ablation}. \textit{Ours\_clip} retains only the segment-level search and two inspection tools, emphasizing reasoning via local grounding. \textit{Ours\_clip\_t} further disables the \textit{Segment Visual Search} from \textit{Ours\_clip}, focusing exclusively on textual retrieval at segment level. \textit{Ours\_visual} removes audio transcripts and speaker annotations while keeping all other components, allowing us to assess the effectiveness of purely visual-structural reasoning. \textit{Ours\_trans} removes speaker identities by the diarization step and constructs the hierarchical database without any speaker-related information.

\begin{table}[ht]
  \small
  \caption{Ablation study on LVBench. All variants use 0.67 fps in captioning. ``Spk.'', ``Trans.'', ``Hier.'', and ``Vis emb.'' denote \textit{Speaker}, \textit{Transcript}, \textit{Hierarchical}, \textit{Visual embed}, respectively.}
  \label{tab:ablation}
  \centering
  \begin{tabular}{lcccc|c}
    \toprule
    Methods & Spk. & Trans. & Hier. & Vis emb.  & Acc.(\%) \\
    \midrule
    Ours\_clip\_t & \checkmark &\checkmark & & & 72.1 \\    
    Ours\_clip & \checkmark &\checkmark  & & \checkmark &  72.8\\  
    Ours\_visual & & & \checkmark & \checkmark & 71.7 \\
    Ours\_trans  & & \checkmark & \checkmark & \checkmark & 75.7 \\
    \midrule
    Ours & \checkmark &\checkmark & \checkmark & \checkmark & 81.0 \\
    \bottomrule
  \end{tabular}
\end{table}

\begin{figure*}[ht] 
  \centering
  \includegraphics[width=\textwidth]{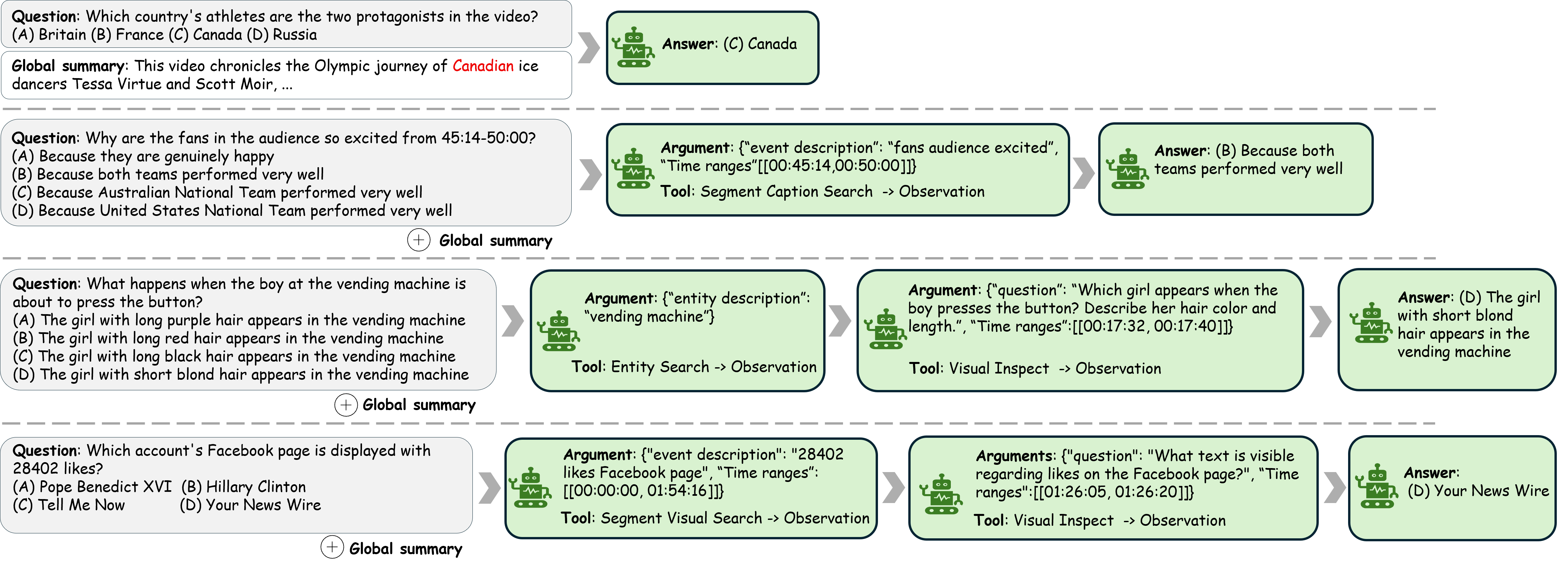}
  \caption{Case study on different reasoning chains with tool calls.}
  \label{fig:case_study}
\end{figure*}

As shown in Table~\ref{tab:ablation}, removing the hierarchical organization in \textit{Ours\_clip} leads to a substantial performance drop compared to \textit{Ours}, with accuracy dropping from $81.0$ to $72.8$. This underscores the central role of hierarchical indexing in aggregating multi-granular evidence and supporting effective cross‑segment reasoning. The audio modality also contributes significantly to overall performance. In \textit{Ours\_visual}, the accuracy drops to $71.7$ due to the loss of transcripts and speaker information that not only provide complementary context beyond visual cues but also enhance entity consistency. Indeed, transcripts alone provide meaningful semantic enrichment: \textit{Ours\_trans} improves by 4\% over \textit{Ours\_visual}, indicating that textual audio cues facilitate richer reasoning. Furthermore, incorporating speaker identity yields an additional 5.3\% accuracy gain when comparing \textit{Ours} with \textit{Ours\_trans}. These results emphasize the effectiveness of both transcripts and speaker labels. Although most questions in LVBench focus on visual content, the audio stream nonetheless strengthens entity coherence, thereby improving semantic consistency within the hierarchical structure. Lastly, \textit{Ours\_clip\_t} performs slightly worse than \textit{Ours\_clip}, because it lacks visual embedding-based search, which offers additional grounding capabilities. All these results demonstrate that these components contribute uniquely to the system’s overall effectiveness.

\subsubsection{Analysis on Six Categories}
\begin{figure}[ht]
  \centering
   \includegraphics[width=0.9\linewidth]{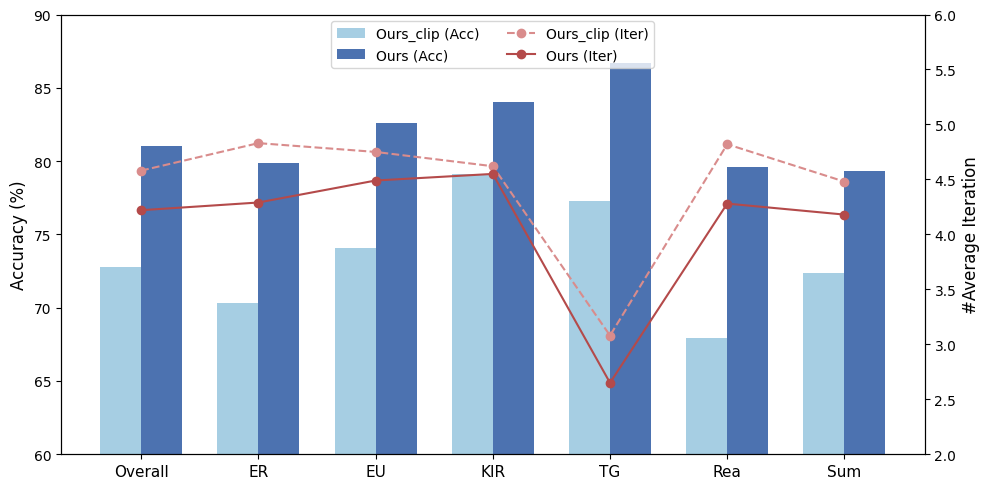}
   \caption{Comparison of accuracy and efficiency on six categories. ``Acc" and ``iter" denote accuracy and the average number of reasoning iterations, respectively.}
   \label{fig:efficiency}
\end{figure}
To further assess the effectiveness of our method across different query types, we compare \textit{Ours} with \textit{Ours\_clip} in terms of accuracy and the average number of reasoning iterations required to answer queries across all six categories. As shown in Figure~\ref{fig:efficiency}, our approach consistently achieves higher accuracy with fewer iterations compared to the non-hierarchical baseline, with particularly notable gains in entity recognition and reasoning. This improvement is expected, as hierarchical indexing provides readily accessible information at multiple granularities, enabling the agent to retrieve relevant context more efficiently and thereby reducing reasoning steps. It is worth noting that the summarization category in LVBench often involves fine-grained, step-level queries (e.g., \textit{“What does Mandy do after she stands in front of the judges’ table?”}), which require multi-turn reasoning rather than simple scene summarization.

\subsection{Qualitative Analysis}
We present several qualitative examples in Figure~\ref{fig:case_study} to illustrate the agent’s adaptive reasoning behavior across diverse question types. For global queries requiring holistic understanding (e.g., \textit{the athletes’ nationality}), the agent retrieves the answer directly from the global summary without invoking additional tools. In contrast, for queries involving specific entities (e.g., \textit{the vending machine} or \textit{the boy}), the \textit{Entity Search} tool is first employed to localize the relevant visual context, followed by targeted inspection to infer detailed attributes or actions. For queries demanding fine-grained visual cues (e.g., \textit{numerical content on a webpage}) that may not be captured in captions, the \textit{Segment Visual Search} tool effectively supplements missing information. Finally, when the query specifies an explicit temporal scope and sufficient contextual evidence is available, the \textit{Visual Inspect} tool is utilized to verify and refine the final answer. These examples highlight the model’s capability to dynamically compose tool chains according to question type and evidence distribution across modalities, demonstrating robust adaptability in complex reasoning scenarios.
\section{Conclusion}
In this work, we address the challenges of long video understanding with a unified framework that combines offline hierarchical video indexing and agentic multi-granularity retrieval. Our approach organizes video content into four levels, global, scene, segment, and entity, while incorporating audiovisual entity cohesion to maintain semantic consistency over extended temporal spans. Extensive experiments show that our method consistently outperforms state-of-the-art baselines, demonstrating the effectiveness of hierarchical indexing and audiovisual entity consolidation. These results highlight the promise of structured, multi-level representations for advancing long video comprehension and motivate further research in this direction.

{
    \small
    \bibliographystyle{ieeenat_fullname}
    \bibliography{main}
}

\clearpage
\setcounter{page}{1}
\maketitlesupplementary
\appendix

The supplemental material contains additional implementation details as well as more results and discussions.

\section{Entity-Centric Re-Captioning}
\label{supp:entity-centric}
After consolidation, each canonical entity $\tilde{e}_j$ is associated with a global description and a set of linked segments $Q_j$. During retrieval, incorporating all linked segments of top-$K_1$ entities can be computationally expensive and may introduce query-irrelevant noise. Such noise can weaken embedding-based matching; while LLM-based re-ranking may mitigate this issue, it incurs higher computational cost. To address this, we further introduce an \textit{entity-centric re-captioning} process during offline construction. For each linked segment \(i\) and entity $\tilde{e}_j$, we generate a focused description \(\tilde{C}^t_{i,j}\) using an LLM that summarizes the entity's appearance, actions, and events within the segment, while excluding irrelevant context. The final entity database contains both canonical entities and fine-grained entity-segment descriptions: $\tilde{\mathcal{E}}=\{\tilde{\mathcal{E}}_g \, ; \, \tilde{\mathcal{E}}_e\}$, where $\tilde{\mathcal{E}}_e=\{\tilde{C}^t_{i,j}|j=1,2,\ldots,J, i \in Q_j\}$. During retrieval, we first match entities in the embedding space of $\tilde{\mathcal{E}}_g$, then re-rank linked segments using similarity between the query and \(\tilde{C}^t_{i,j}\), selecting the top-$K_2$ segments for precise grounding. This design balances retrieval precision and computational efficiency, avoiding excessive LLM overhead. In our implementation, $K_1$ and $K_2$ are set to 20 and 16, respectively.

\section{Multi-Granularity Tools}
\label{supp:tools}
Here we present details of the whole tool set denoted by
\begin{equation}
    \mathcal{T} = \{ T_{\text{scene}}, T_{\text{caption}}, T_{\text{visual}}, T_{\text{entity}}, T_{\text{inspect}} \}.
\end{equation}

\noindent\textbf{Global Scene Browse.} 
This tool $T_{\text{scene}}$ supports coarse-grained navigation and scene localization along the video timeline. Given a user query $q$ and the scene collection $D = \tilde{\mathcal{S}}$, it identifies and summarizes the most relevant scenes with an LLM, returning their storyline and corresponding timestamps $\tau'$. The agent tends to invoke this tool for complex or ambiguous queries involving multiple events or temporal dependencies.

\noindent\textbf{Segment Caption Search.}
This tool $T_{\text{caption}}$ performs fine-grained text-based retrieval within specified temporal ranges. Given the user query $q$ and the segment database $D = \tilde{\mathcal{C}}$, the tool retrieves the most semantically relevant segment descriptions $r$ along with their associated time spans $\tau'$. This is achieved through cosine similarity matching between the query embedding and pre-computed caption embeddings for all video segments, ensuring efficient and accurate retrieval of localized content. 

\noindent\textbf{Segment Visual Search.}
To capture visual cues that may be overlooked in textual descriptions, the \textit{Segment Visual Search} tool $T_{\text{visual}}$ complements $T_{\text{caption}}$. While the latter relies on text embeddings, $T_{\text{visual}}$ leverages cross-modal embeddings generated by the UNITE framework~\cite{unite}. This design enables retrieval driven by rich visual semantics aligned with the query, ensuring that visually salient details are incorporated into the search process.

\noindent\textbf{Entity Search.}
This tool $T_{\text{entity}}$ supports high-level, entity-centric retrieval across large temporal ranges. Given an entity-related query $q$, the tool first retrieves the top-$K_1$ most relevant entities from database $\tilde{\mathcal{E}_g}$ based on their descriptions in the pre-computed embedding space. For all segments linked to these entities, it then performs a second-stage reranking to select the top-$K_2$ most relevant segments from the entity-centric database $\tilde{\mathcal{E}_e}$, using the same query. Finally, the tool applies entity-aware re-captioning to the retrieved segments, generating a coherent response $r$ enriched with precise timestamps $\tau'$. 

\noindent\textbf{Inspection Tool.}
This tool $T_{\text{inspect}}$ provides fine-grained temporal inspection to support detailed reasoning. It consists of two complementary modules: \textit{Clip Caption Inspect} ($T_{\text{inspect}}^{\text{tex}}$) and \textit{Visual Inspect} ($T_{\text{inspect}}^{\text{vis}}$). The $T_{\text{inspect}}^{\text{tex}}$ examines coarse textual descriptions to determine what occurred during a specified time span. For example, for a query such as \textit{“What does the protagonist do after he jumps down the stairs?”}, the tool inspects subsequent time ranges to identify the protagonist’s next actions after locating the event \textit{“he jumps down the stairs”} in previous iteration. The $T_{\text{inspect}}^{\text{vis}}$ tool leverages a VLM to perform precise visual verification within a given time range. Due to frame limits of the VLM, this inspection focuses on short intervals, ensuring accurate visual grounding for fine-grained queries.

\begin{figure*}[ht] 
  \centering
  \includegraphics[width=\textwidth]{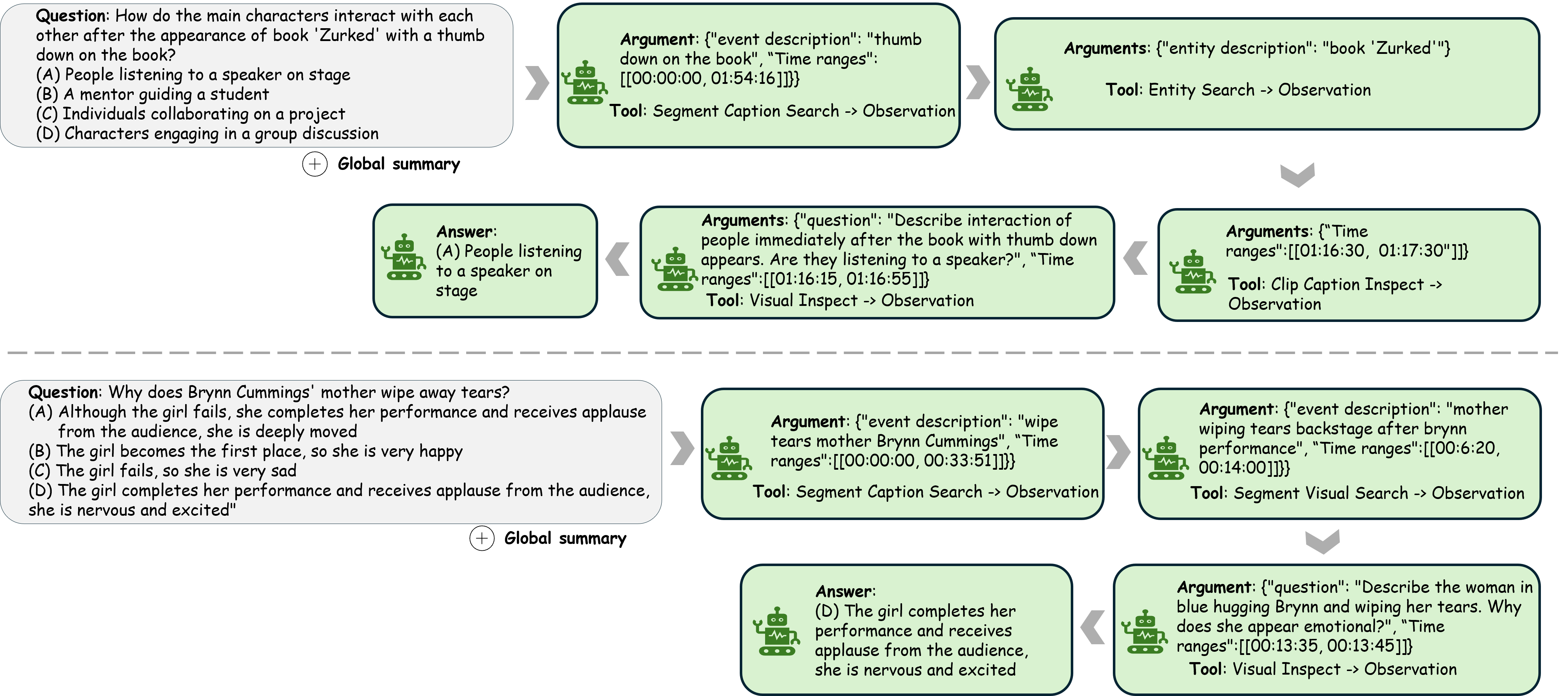}
  \caption{Cases with multiple reasoning steps.}
  \label{fig:case_2}
\end{figure*}

\section{Other Technical Details}
We leverage agglomerative clustering with a threshold of 0.4 for entity clustering stage. A chunk size of 24 with an overlap of 3 is employed for scene segmentation. 
Due to the API frame limit, the \textit{Ours (2 fps)} variant generates captions for two 15-second sub-segments, each sampled with 30 frames, and concatenates them into a single description.

\section{Additional Case Study}
As shown in Figure~\ref{fig:case_2}, we present more complex question-answer cases that involve multiple reasoning steps and tool calls, further demonstrating the effectiveness of our proposed pipeline. For the interaction question involving \textit{the appearance of the book “Zurked” with a thumbs-down}, the agent first performs a coarse segment-level caption search, then progressively narrows the temporal window via entity search. After validating the candidate clips with caption inspection in a long time range, it applies fine-grained visual inspection to extract the correct interaction outcome. For the reasoning-oriented question \textit{“Why does Brynn Cummings’ mother wipe away tears?”}, the agent again initiates with segment caption search tool, but due to incomplete textual matches, it escalates to segment-level visual search to gather more reliable evidence. Once the event is localized, the agent conducts a targeted visual inspection to infer the emotional cause behind the mother’s reaction.

\section{Efficiency}
Our semantic-consistent hierarchy enables more efficient navigation, achieving higher accuracy with fewer reasoning steps and less runtime compared with DVD, as shown in Table \ref{tab:efficiency}.

\begin{table}[ht]
  \small
  \caption{Comparison of average number of iterations and runtime (second per query).}
  \label{tab:efficiency}
  \centering
  \begin{tabular}{lcc}
    \toprule
    Methods & Iteration & Runtime (s)  \\
    \midrule
    DVD & 7.6 & 151.0 \\    
    Ours & 4.2 & 98.7 \\  
    \bottomrule
  \end{tabular}
\end{table}

\section{Proprietary models}
Our API versions are GPT-4.1 (2025-04-14) and o3 (2025-04-16). Variance of three runs on LVBench is 0.149, which demonstrates the robustness of our method. Furthermore, Our method achieves an competitive accuracy of 75.8\% on LVBench using open-source models (DeepSeek-R1-0528 for reasoning + Qwen3-VL-32B-Instruct for visual inspection). 

\section{Prompt for the Planner}
We present the planner’s prompt for agentic search in Table~\ref{tab:agent_prompt}. This prompt guides the planner in selecting the most appropriate tools to search from the hierarchical database and determining what information to request at each reasoning step, thereby enabling systematic information gathering and progressively moving toward the final answer. 

\begin{table*}[t]
    \centering
    \small 
    \renewcommand{\arraystretch}{1.25} 
    
    \caption{The agentic search prompt structure. The prompt is divided into four distinct sections: goal, tools, tool preferences and hints.}
    \label{tab:agent_prompt}
    
    \begin{tabular}{p{0.12\linewidth} p{0.85\linewidth}}
        \toprule
        \textbf{Module} & \textbf{Prompt Content} \\
        \midrule

        \textbf{GOAL} & 
        You will be given a set of tools to assist you exploring the video, understanding it and reasoning the answer. \newline
        Please follow the THINK $\rightarrow$ ACT $\rightarrow$ OBSERVE loop: \newline
        $\bullet$ THOUGHT: Reason step-by-step about what question to ask and which tool to call next. \newline
        $\bullet$ ACTION: Call exactly one tool that moves you closer to the final answer. \newline
        $\bullet$ OBSERVATION: Summarize the tool call's output. \newline
        Continue the loop until the user's query is fully resolved, then end your turn with the final answer.
        \\ \midrule
        
        \textbf{TOOLS} & 
        \small 
        Here are tools you can use to reveal your reasoning process whenever the provided information is insufficient: \newline
        $\bullet$ \textbf{global\_scene\_browse\_tool}: for scene-related query to explore scenarios, temporal orders, and contextual structure in a rough manner without precise details (e.g., first appearance, second song, third collision). \newline
        $\bullet$ \textbf{entity\_search\_tool}: for entity-related information retrieval, finding important subjects involved in events. \newline
        $\bullet$ \textbf{clip\_caption\_search\_wtime\_tool}: to search from rough captions and audio transcriptions of local clips within a list of time ranges related to a query. If you want to search from the whole video, use the whole video time range. \newline
        $\bullet$ \textbf{clip\_visual\_search\_wtime\_tool}: to search from visual features of local clips within a list of time ranges (list[tuple[HH:MM:SS, HH:MM:SS]]) related to a query. If you want to search from the whole video, use the whole video time range. This tool may provide more detailed information as a supplement to \textbf{clip\_caption\_search\_wtime\_tool}. \newline
        $\bullet$ \textbf{clip\_caption\_inspect\_tool}: to extract rough captions and audio transcriptions of local clips within any list of time ranges (list[tuple[HH:MM:SS, HH:MM:SS]]). This tool is suitable for further inspecting what happened in a time range before or after some events happened. \newline
        $\bullet$ \textbf{visual\_inspect\_tool}: to extract details from local visual clips within a narrow list of time ranges that covers less than 50 seconds to answer a question or retrieve query-related details. When the time ranges cover over 50 seconds, first use \textbf{clip\_caption\_inspect\_tool} to get a rough context. \newline
        $\bullet$ \textbf{finish}: Once you believe you have found the answer, you can call the \textbf{finish} tool with an answer.
        \\ \midrule
        
        \textbf{TOOL PREFERENCES} & 
        $\bullet$ When no context is given, call \textbf{global\_scene\_browse\_tool} for scene-related queries to get an overview of related context and timelines, \newline
        or call \textbf{clip\_caption\_search\_wtime\_tool} for event-related queries to get a rough context and timelines, \newline
        or call \textbf{entity\_search\_tool} for entity-related queries. \newline
        $\bullet$ When you cannot locate the needed context from scene or entity tools, use \textbf{clip\_caption\_search\_wtime\_tool} or \textbf{clip\_visual\_search\_wtime\_tool} to expand your search. \newline
        $\bullet$ If the retrieved material by \textbf{clip\_caption\_search\_wtime\_tool} lacks relevant contexts, further call \textbf{clip\_visual\_search\_wtime\_tool} for more fine-grained search. \newline
        $\bullet$ If the retrieved material lacks precise, question-relevant detail (e.g., an unknown name, count) or you are uncertain of an answer after searching, call \textbf{clip\_caption\_inspect\_tool} \newline
        or inspect frames with \textbf{visual\_inspect\_tool} with a list of time ranges to take a closer look. \newline
        $\bullet$ After locating an answer in the script, always make a \textbf{CONFIRM} with \textbf{visual\_inspect\_tool} query. 
        \\ \midrule
        
        \textbf{HINTS} & 
        $\bullet$ Before giving the final answer, confirm critical visual or numeric facts with \textbf{visual\_inspect\_tool}. \newline
        $\bullet$ If you call \textbf{clip\_caption\_search\_wtime\_tool} in three consecutive times but still cannot find useful information to answer the question, please try \textbf{clip\_visual\_search\_wtime\_tool} to get more detailed information. \newline
        $\bullet$ You have at most 10 iterations of THINK$\rightarrow$ACT$\rightarrow$OBSERVE; plan strategically. Avoid redundant information retrieval steps. \newline
        $\bullet$ To make a good plan to questions that need complex reasoning, sometimes you need to first ask some other related contexts instead of directly asking the target question, \newline
        $\bullet$ For questions that need counting the number of times an event occurs over time, call \textbf{global\_scene\_browse\_tool} first. \newline
        If you are uncertain about its answer, please search for related events or subjects without counting first to find all related information, then do the counting based on observations. \newline
        $\bullet$ Your final answer must be concise and directly address the question.

        \\ \bottomrule
        
    \end{tabular}
\end{table*}

{
    \small
}

\end{document}